\documentclass[sigconf,natbib=true,anonymous=false]{acmart}
\renewcommand\footnotetextcopyrightpermission[1]{} 
\usepackage{CJKutf8}
\usepackage{multirow}
\usepackage{todonotes}
\usepackage{subfigure} 
\usepackage[ruled,vlined]{algorithm2e} 
\usepackage{algorithmic}
\usepackage[capitalize,noabbrev]{cleveref}
\usepackage{colortbl}
\usepackage{xspace}

\newcommand{\ours}{\textit{MCSCSet}\xspace}
\newcommand{\eg}{\textit{e.g\@.}}
\newcommand{\ie}{\textit{i.e\@.}}

\newcommand{\chinese}[1]{\begin{CJK*}{UTF8}{gkai}{#1}\end{CJK*}}
\usepackage{adjustbox}
\newcommand{\red}[1]{{\color{red} #1}}
\DeclareMathOperator*{\argmax}{argmax}

\begin{document}

\title{MCSCSet: A Specialist-annotated Dataset for Medical-domain Chinese Spelling Correction}

\author{Wangjie Jiang\textsuperscript{$\dagger$}, 
Zhihao Ye\textsuperscript{$\ddagger$}, 
Zijing Ou\textsuperscript{$\spadesuit$}, 
Ruihui Zhao\textsuperscript{$\ddagger$}, 
Jianguang Zheng\textsuperscript{$\ddagger$}, 
Yi Liu\textsuperscript{$\ddagger$}, \\ 
Siheng Li\textsuperscript{$\dagger$}, 
Bang Liu\textsuperscript{$\clubsuit$}, 
Yujiu Yang\textsuperscript{$\dagger$} and Yefeng Zheng\textsuperscript{$\ddagger$}}

\affiliation{%
  \institution{\textsuperscript{$\dagger$}Tsinghua Shenzhen International Graduate School, Tsinghua University\state{Shenzhen} \country{China}\\ \textsuperscript{$\ddagger$}Tencent Jarvis Lab \state{Shenzhen} \country{China}\\ \textsuperscript{$\spadesuit$}Sun Yat-sen University \state{Guangzhou} \country{China}\\ \textsuperscript{$\clubsuit$}Université de Montréal Mila \& CIFAR\state{Québec} \country{Canada}}
}
\email{{jwj20,lisiheng21}@mails.tsinghua.edu.cn, {evanzhye,zacharyzhao,jaxzheng,yefengzheng}@tencent.com}
 \email{{zijingou.mail,97liuyi}@gmail.com, bang.liu@umontreal.ca, yang.yujiu@sz.tsinghua.edu.cn}

\renewcommand{\shortauthors}{Wangjie Jiang et al.}

\begin{abstract}
Chinese Spelling Correction (CSC) is gaining increasing attention due to its promise of automatically detecting and correcting spelling errors in Chinese texts. Despite its extensive use in many applications, like search engines and optical character recognition systems, little has been explored in medical scenarios in which complex and uncommon medical entities are easily misspelled. Correcting the misspellings of medical entities is arguably more difficult than those in the open domain due to its requirements of specific domain knowledge. In this work, we define the task of Medical-domain Chinese Spelling Correction and propose \ours, a large-scale specialist-annotated dataset that contains about \textit{200k} samples. In contrast to the existing open-domain CSC datasets, \ours involves: \textit{i)} extensive real-world medical queries collected from Tencent Yidian, \textit{ii)} corresponding misspelled sentences manually annotated by medical specialists. To ensure automated dataset curation, \ours further offers a medical confusion set consisting of the commonly misspelled characters of given Chinese medical terms. This enables one to create the medical misspelling dataset automatically. Extensive empirical studies have shown significant performance gaps between the open-domain and medical-domain spelling correction, highlighting the need to develop high-quality datasets that allow for Chinese spelling correction in specific domains. Moreover, our work benchmarks several representative Chinese spelling correction models, establishing baselines for future work. 
\end{abstract}

\maketitle

\section{Introduction} \label{sec_introduction}
Misspelled characters frequently occur in hand-crafted Chinese sentences, easily leading to a wrong understanding of these sentences. To this end, we need a corrector to automatically detect and correct spelling mistakes in the text. The task of Chinese Spelling Correction (CSC) is to design such a corrector to correct spelling errors, which plays a vital role in various Natural Language Processing (NLP) applications such as search engine~\cite{martins2004spelling} and optical character recognition system~\cite{afli2016using}. To achieve the goal of efficient error correction, previous work has mainly focused on designing advanced error correction models \cite{cheng2020spellgcn,zhang2020spelling,hong2019faspell,yeh2014chinese} and establishing canonical benchmark spelling correction corpora~\cite{wu2013chinese,yu2014chinese,tseng2015introduction,liu2021plome}. For example, a well-known open-domain spelling correction corpus, \textit{SIGHAN-15} \cite{tseng2015introduction}, is a Chinese spelling correction corpus collected from a computer-based Test of Chinese as a Foreign Language (TOCFL). Although these models and benchmark datasets provide people with high-quality spelling error correction services in the open domain, their effectivenss is reduced significantly in some specific domains, such as the medical domain. The reason is that open-domain corpora do not contain complex medical terms, and the spelling of medical terms requires specialized domain knowledge that ordinary people usually lack~\cite{moon2007ontology,zhao2017hybrid}.

Additionally, Chinese spelling correction for medical terms plays a crucial role in promoting the standardization and healthy development of the medical field~\cite{zhou2015context}. Indeed, Chinese spelling correctors may improve the quality of medical application services, especially medical entity search systems, by automatically correcting medical terms with misspellings. Specifically, we get incorrect answers when using the medical entity query system to query medical terms with misspellings, leading to user misunderstandings and even severe medical malpractice. For example, when a doctor's hand-crafted electronic medical record contains misspellings for a malignant disease, a patient queries the term and may get results that misidentify themselves as having another illness, delaying patient care, and affecting a healthy doctor-patient relationship. This indicates that the spelling error in the medical field, especially in the medical entity query scenario, need to be corrected and resolved urgently~\cite{yeh2015chinese}. Therefore, we need to find an effective way to correct spelling mistakes in the medical domain.

\begin{table*}[!t]
\centering
\caption{Performance of a well-trained open-domain BERT-based CSC model on detection-level and correction-level tasks. Specifically, the model is first pre-trained on large-scale automatically-generated data \cite{li-etal-2021-exploration} and then fine-tuned on \textit{SIGHAN-15} \cite{tseng2015introduction} . We report the model's performances on test sets of \textit{SIGHAN-15} and the proposed \ours, respectively.}
\setlength{\tabcolsep}{7 mm}{
\begin{tabular}{c|ccc|ccc}
\toprule
\multicolumn{1}{c|}{\multirow{2}{*}{Test Set}}   & \multicolumn{3}{c|}{Detection-level} & \multicolumn{3}{c}{Correction-level}   \\
\cline { 2 - 7 } &  Prec. ($\%$) & Rec. ($\%$)& F1 ($\%$) & Prec. ($\%$) & Rec. ($\%$) & F1 ($\%$) \\
\midrule
\textit{SIGHAN-15} & 79.06 & 83.73 & 81.33 & 77.31 & 81.89 & 79.53  \\
\ours  &  43.83 &  38.94 &  41.24 &  28.58 &  25.38 &  26.89  \\
\bottomrule
\end{tabular}
}
\label{tab:openmodel_on_CMQSC}
\end{table*}

To achieve this goal, a straightforward method is to directly apply advanced methods~\cite{liu2021plome,huang2021phmospell,guo2021global,xu-etal-2021-read,zhang2021correcting} in open-domain CSC to medical-domain CSC.
However, such a method is likely to fail on the medical CSC task due to the offset of the corresponding domain knowledge. 
To verify this, we choose an advanced BERT-based CSC model \cite{li-etal-2021-exploration}, which is first pre-trained on large-scale automatically-generated CSC data and then fine-tuned on \textit{SIGHAN-15}. 
Then we validate the model on the test sets of \textit{SIGHAN-15} and our proposed medical-domain dataset in this paper.
The experimental results are shown in \cref{tab:openmodel_on_CMQSC}, and it can be seen that such a naive method shows a significant performance gap between in-domain and out-of-domain experiments. 
We conjecture that this is because the distribution of spelling errors differs significantly between an open domain and a specific domain. 
For instance, in Chinese medical texts, the vast majority of spelling errors occur in those complex and uncommon medical entities, which rarely occur in the open-domain Chinese texts, \eg, \textit{SIGHAN-15}, which is collected from TOCFL. 
In particular, we summarize the errors of medical terms into five categories of which the phonological errors and the visual errors belong to spelling errors, and show their corresponding examples in Table \ref{tab:type_of_spelling_errors}. 
We can observe from the table that the errors in the medical domain are not common in the open domain, which highlights the need to develop high-quality datasets that allow for medical-domain Chinese spelling correction. 

\begin{CJK*}{UTF8}{gkai}
\begin{table}[t]
\centering
\small
\caption{Examples of typical Chinese medical entity errors, which can be mainly divided into five categories: i) phonological errors, ii) visual errors, iii) order-confused errors; iv) repeated characters, and v) missing characters. Among the five categories, phonological and visual errors belong to spelling errors, which are the focus of our study. Erroneous characters are marked in {red}, and the corresponding phonics are given in brackets.}
\label{tab:type_of_spelling_errors}
\vspace{2pt}
\begin{tabular}{lll}
\rowcolor{black} \textcolor{white}{Type} & \textcolor{white}{Sentence} & \textcolor{white}{Correction}  \\
\hline
{\cellcolor{gray!10}} & \cellcolor{gray!10}如何\red{闭(bi)孕} & \cellcolor{gray!10}避(bi)孕 \\\multirow{-2}*{\cellcolor{gray!10}{Phonological}}
& \cellcolor{gray!10}how to \red{close pregnancy} & \cellcolor{gray!10}contraception \\
\hline
\multirow{2}*{Visual} & 胰岛素应该用\red{水箱}储存吗 & 冰箱 \\
& should insulin stored in \red{water tank} & refrigerator \\
\hline
\cellcolor{gray!10} & \cellcolor{gray!10}如何处理\red{蜂蜜}蛰伤 & \cellcolor{gray!10}蜜蜂 \\
\multirow{-2}*{\cellcolor{gray!10}Order-confused}& \cellcolor{gray!10}how to deal with \red{honey} stings & \cellcolor{gray!10}bee \\
\hline
\multirow{2}*{Redundant} & \red{天花粉}的症状 & 天花 \\
& symptoms of \red{trichosanthin} & smallpox \\
\hline
\cellcolor{gray!10} & \cellcolor{gray!10}糖尿病患者能服用\red{葡萄}吗 & \cellcolor{gray!10}葡萄糖 \\
\multirow{-2}*{\cellcolor{gray!10}Missing}& \cellcolor{gray!10}can diabetics take \red{grapes} & \cellcolor{gray!10}glucose \\
\hline

\end{tabular}
\end{table}
\end{CJK*}

Here, we highlight the challenges of building a large-scale Chinese spelling correction benchmark dataset in the medical domain as follows: 

\noindent \textbf{(C1)} \textbf{Difficulty to Collect Real Data:} To be able to provide the service of medical entity error correction in real application scenarios, annotated datasets must come from real medical scenarios and contain common error-prone medical entities among the hundreds of millions of queries generated by real-world applications.

\noindent \textbf{(C2)} \textbf{High Demand of Medical Knowledge:} To produce a high-quality medical term (or entity) spelling correction corpus, annotators are required to master specific medical knowledge and maintain high correction quality, which is a challenging and time-consuming task.

To address the above challenges, in this paper, we present \textit{Medical Chinese Spelling Correction Dataset (\ours)}, a large-scale and specialist-annotated dataset for Chinese spelling correction in the medical domain. Notably, we collect a large-scale query log dataset from a real-world medical application named Tencent Yidian\footnote{\href{https://baike.qq.com/}{https://baike.qq.com/}} and construct a manually annotated dataset with about \textit{200k} samples, in which each sample consists of a correct medical query and its corresponding wrong medical query with spelling errors. \ours also provides a medical confusion set, consisting of a large number of error-prone characters from Chinese medical terminologies, each with its corresponding erroneous characters.
This enables potential researchers or practitioners to generate new medical-domain CSC datasets based on their specific needs by simply replacing the medical entities with misspelled characters defined in the confusion set.
To distinguish from the open-domain CSC, we further provide a formal definition of the medical-domain Chinese spelling correction task, mainly focusing on the spelling error correction for medical entities. 
Moreover, our work benchmarks several Chinese spelling correction models for future comparisons. 
Overall, the following components summarize our major contributions:

\begin{itemize}
    \item \textbf{Practical Task Definition of Medical-domain Chinese Spelling Correction:} We formally define the Chinese spelling correction task in the medical domain for the first time, which applies to all tasks involving user input such as search, question answering, and translation.
    \item \textbf{First CSC Dataset for Medical Domain:} We provide the first Chinese medical spelling correction dataset from the large-scale healthcare encyclopedia software Tencent Yidian, based on the annotation of medical specialists.
    \item \textbf{Rich Medical Confusion Set:} We present a corresponding medical confusion set, which consists of abundant error-prone medical entities. This allows great flexibility for future usage since one could exploit it to construct a new dataset.
    \item \textbf{Rigorous Medical-domain CSC Benchmarking:} We benchmark four representative Chinese spelling correction models, which verify the quality of the proposed \ours dataset and provide reproducible comparisons for future studies.
\end{itemize}
\textbf{Paper Organizations.} \cref{sec_related_work} presents background and related work on Chinese spelling correction, including previous CSC algorithms, datasets, and benchmarks. In \cref{sec_task_definition}, we present the definition of the problem of medical-domain Chinese spelling correction. In \cref{sec_cmqsc} we provide details on the construction process of the \ours dataset and present some statistical analysis. \cref{sec_experiment} provides specifics of benchmarking representative CSC algorithms, implementation details and experimental results. Lastly, \cref{sec_conclusion} discusses and concludes the paper.

\begin{table*}[t]
\centering
\caption{The comparison of \ours and existing CSC datasets. }
\setlength{\tabcolsep}{5 mm}{
\begin{tabular}{cccccc}
\rowcolor{black} \textcolor{white}{Dataset} & \textcolor{white}{\# Sentences} & \textcolor{white}{Avg. length} & \textcolor{white}{\# Error types} & \textcolor{white}{Domain} & \textcolor{white}{Annotation} \\ 
\hline 
\textit{SIGHAN-13} \cite{wu-etal-2013-chinese} & 1,700 & 60.9 & 2 & Open & Human-annotation \\
\textit{SIGHAN-14} \cite{yu2014chinese} & 4,463 & 49.7 & 2 & Open & Human-annotation \\
\textit{SIGHAN-15} \cite{tseng2015introduction} & 3,438 & 31.1 & 2 & Open & Human-annotation \\
\textit{OCRSet} \cite{hong2019faspell} & 4,575 & 10.2 & 2 & Open & Automatically-generated \\
\textit{HybridSet} \cite{wang2018hybrid} & 271,329 & 42.5 & 2 & Open & Automatically-generated \\
\cellcolor{gray!10}\textbf{\ours} & \cellcolor{gray!10}\textbf{199,877} & \cellcolor{gray!10}\textbf{10.9} & \cellcolor{gray!10}\textbf{5} & \cellcolor{gray!10}\textbf{Medical} & \cellcolor{gray!10}\textbf{Specialist-annotation} \\
\hline
\end{tabular}}
\label{tab:statistics_of_CSCdataset}
\end{table*}

\begin{CJK*}{UTF8}{gkai}
\begin{table}[!t]
\centering
\caption{A training sample of Chinese medical query spelling correction task. Here, the misspelled medical entity ``拨智尺'' (dial the wisdom ruler) should be corrected to ``拔智齿'' (extract the wisdom tooth).}
\setlength{\tabcolsep}{1mm}{
\begin{tabular}{l}
\toprule
Wrong query: \red{\underline{拨}}智\red{\underline{尺}}的过程。   \\
Translation: The process of \red{\underline{dial}} the wisdom \red{\underline{ruler}}.   \\
\midrule
Output query: \red{\underline{拔}}智\red{\underline{齿}}的过程。  \\
Translation: The process of wisdom \red{\underline{tooth}} \red{\underline{extraction}}.   \\
\bottomrule
\end{tabular}
}
\label{tab:error_example}
\end{table}
\end{CJK*}

\section{Related Work} \label{sec_related_work}

\noindent \textbf{Chinese Spelling Correction.}
Chinese Spelling Correction (CSC) is a challenging task in Natural Language Processing (NLP) and plays an important part in various real-world applications, such as search engine \cite{martins2004spelling, gao2010large, duan2018error}, optical character recognition \cite{wang2018hybrid, mokhtar2018ocr, drobac2020optical} and automatic speech recognition \cite{guo2019spelling, zhang2019investigation, yang2019post, arslan2021detecting}. CSC is similar to the task of Chinese Grammatical Error Correction (CGEC) \cite{lin2015ntou}. The difference between them is that CSC only focuses on Chinese spelling errors that need replacement, while CGEC also includes errors that require deletion and insertion.

\noindent \textbf{Existing Methods for CSC.}
CSC has been an active research topic for many years. Early works followed error detection, correction, candidate generation, and candidate selection. Most of the proposed correction methods \cite{xie2015chinese, liu2013hybrid, yu2014chinese} employed unsupervised n-gram language models and confusion set to detect errors and select candidate replacement characters to fix the errors. With the development of deep learning and especially pre-trained language models (PLM) \cite{qiu2020pre} in recent years, great progress has been achieved in the CSC task. Since the pre-trained task (\ie, masked language model) of BERT is similar to CSC, many approaches utilized BERT \cite{devlin-etal-2019-bert} in their models. Hong et al. \shortcite{hong2019faspell} proposed the FASPell model based on BERT and exploited the phonological and visual similarity information to filter candidate characters. Cheng et al. \shortcite{cheng2020spellgcn} proposed to incorporate the character similarity knowledge into BERT via a Graph Convolution Network (GCN) \cite{KipfW2017gcn}. Zhang et al. \shortcite{zhang2020spelling} proposed the soft-masked BERT model, in which Gated Recurrent Unit (GRU) \cite{gru} was used to detect the erroneous positions and BERT was used to predict correct characters. These methods achieved excellent results in \textit{SIGHAN} \cite{wu-etal-2013-chinese, yu-etal-2014-overview, tseng-etal-2015-introduction}, a series of well-known benchmark datasets in open-domain CSC. 
More recently, some studies \cite{ji2021spellbert, li-etal-2021-exploration, huang2021phmospell, guo2021global, zhang2021correcting} focused on treating CSC as a pre-trained task. Specifically, they utilized the open-domain confusion set to automatically generate a large number of pairs of spelling errors and the corresponding corrections. Models are first pre-trained on the large-scale automatically-generated data and then fine-tuned on the benchmark dataset. Benefiting from the CSC pretraining, better performances have been achieved. Although numerous excellent methods have been proposed to tackle open-domain CSC, few works studied CSC in a specific domain, limiting CSC's wide application.

\noindent \textbf{Existing Datasets for CSC.} There are several public datasets for CSC. Besides the above-mentioned \textit{SIGHAN} series (including \textit{SIGHAN-13} \cite{wu-etal-2013-chinese}, \textit{SIGHAN-14} \cite{yu2014chinese} and \textit{SIGHAN-15} \cite{tseng2015introduction}), \textit{OCRSet} \cite{hong2019faspell} and \textit{HybridSet} \cite{wang2018hybrid} are also commonly used. Table \ref{tab:statistics_of_CSCdataset} shows statistics of the existing CSC datasets and our proposed \ours. We can find that existing public CSC datasets suffer from either limited scale or low quality. The \textit{SIGHAN} series has high quality due to human-annotation but poor in scale. As for \textit{HybridSet}, although the dataset scale is large, the quality is relatively poor because automatically-generated data cannot exactly simulate the distribution of real-world spelling errors.
Besides, these datasets are all collected from open-domain texts and cannot be used in a specific domain such as medicine.

\noindent \textbf{Discussion.}
To tackle the above issues, we take the first step to study CSC in the medical domain.
Specifically, we focus on a real-world medical scenario, \ie, medical search engine, and propose the task of medical-domain Chinese spelling correction. To support this task, we construct a high-quality and large-scale dataset by specialist annotation.

\section{Medical-domain Chinese spelling correction Task} \label{sec_task_definition}

In this section, we introduce our proposed medical-domain Chinese spelling correction, which is a new task on top of open-domain CSC.
Specifically, the formal definition is provided in \cref{definition_medical_csc} and a detailed discussion of the comparison between medical- and open-domain CSC follows immediately after.

\begin{definition}[Medical-domain Chinese Spelling Correction] \label{definition_medical_csc}
\textit{
Given a Chinese medical query $X=\{x_1, x_2,\ldots,x_n\}$ with $n$ characters, the goal of the task is to detect spelling errors on character level and output its corresponding correct sequence $Y=\{y_1, y_2,\ldots,y_m\}$.
Following the general spelling correction task \cite{schutze2008introduction}, the lengths of $X$ and $Y$ are assumed to be equal here, \ie, $n = m$.
We model the task as a conditional generation problem by modeling and maximizing the log likelihood of probability $p_\theta (Y \mid X)$ with parameter $\theta$. Thus, the training objective can be formulated as follows:
\begin{equation}
    \argmax_{\theta} \sum\limits_{(X,Y) \sim \mathcal{D}} \log p_\theta (Y \mid X).
\end{equation}
Where the data point $(X,Y)$ is uniformly sampled from the dataset $\mathcal{D}$, $X$ contains one or more medical entities that include misspelled characters, while the corresponding entities in $Y$ are absolutely correct.
Therefore, the key of medical-domain Chinese spelling correction is to detect and correct misspelled characters in medical entities.
}

\end{definition}

 
To help better understand this task, we give an example in Table \ref{tab:error_example}. 
It can be seen that the input query ``\chinese{拨智尺的过程}'' (\ie, "the process of dialing the wisdom ruler" in English) contains a medical entity ``\chinese{拨智尺}'' that includes two misspelled characters ``\chinese{拨}'' (dial) and “\chinese{尺}'' (ruler).
Our task aims to predict the output query ``\chinese{拔智齿的过程}'' (the process of wisdom tooth extraction), in which the two misspelled characters of the medical entity are corrected to ``\chinese{拔}'' (extract) and ``\chinese{齿}'' (tooth), respectively. 
We can see that ``\chinese{拨}'' (dial) is  visually similar to ``\chinese{拔}'' (extract)  while “\chinese{尺}'' (ruler) is phonetically similar to ``\chinese{齿}'' (tooth), which are error-prone for one without both linguistic knowledge and medical knowledge.


Compared with open-domain CSC, our task is quite different and more difficult. 
First, Chinese medical queries usually contain medical terms that are generally complex and uncommonly used.
Second, spelling errors are more likely to occur in medical entities than other characters.
Third, correcting the spelling errors in medical entities requires medical knowledge in addition to linguistic knowledge.
Therefore, as shown in the \cref{tab:openmodel_on_CMQSC}, directly transferring the model trained from the open domain would significantly degenerate the performance in the medical domain.
To this end, it is necessary to construct a large-scale and high-quality dataset with specialist annotation to facilitate research on our task.


\begin{figure}[t]
\begin{minipage}[t]{1.0\linewidth}
  \centering
  \centerline{\includegraphics[width=8.5cm]{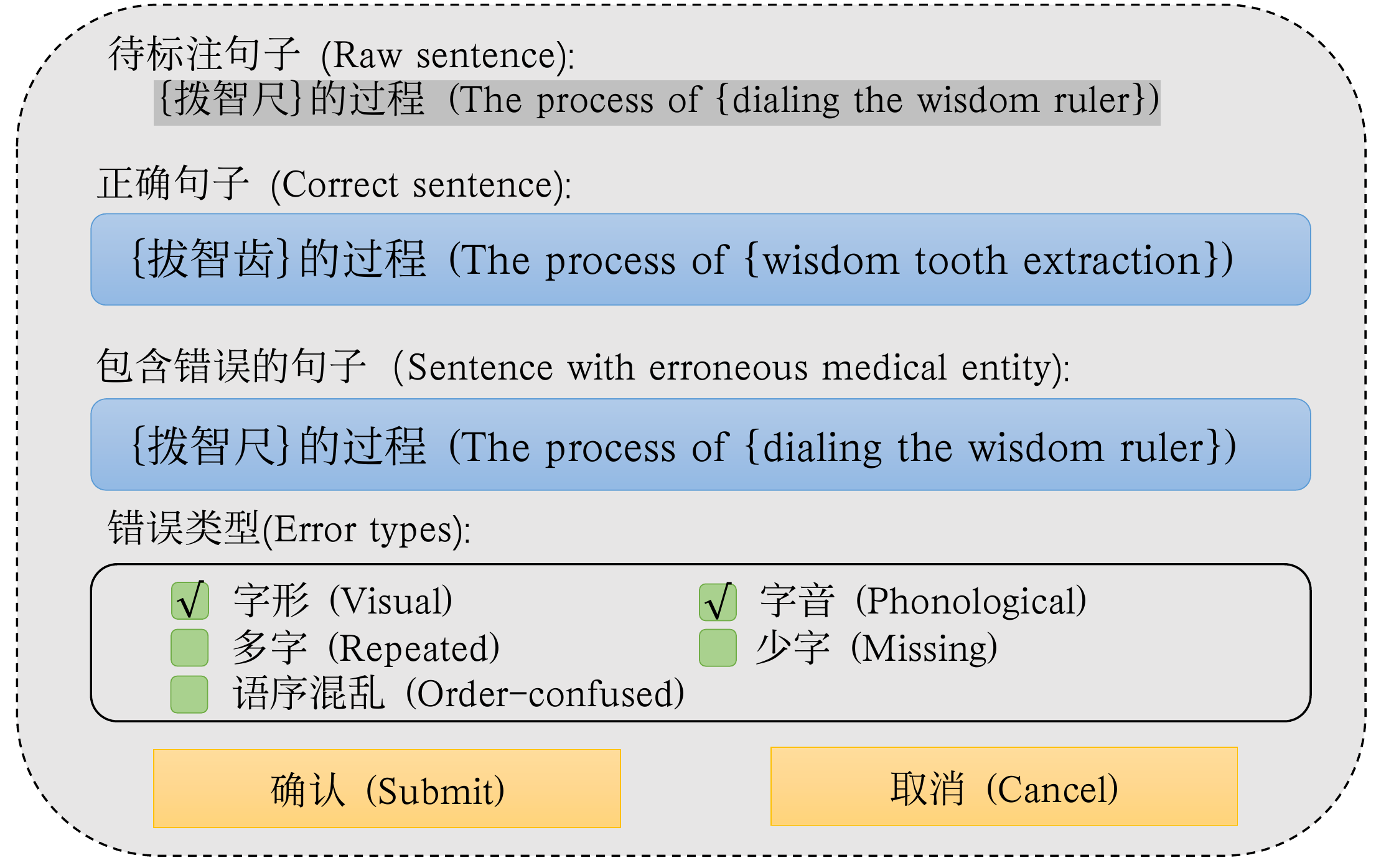}}
\end{minipage}
\caption{Annotation interface.}
\label{fig:interface}
\end{figure}

\section{The Medical-domain Chinese Spelling Correction Dataset} \label{sec_cmqsc}
In this section, we present the medical-domain Chinese spelling correction dataset (\ours).
To help fully understand the property of \ours, we first describe the data collection and annotation steps of medical query errors in \cref{data_collection} and \cref{annotation_process}, respectively. Then the detailed data format and statistics are given in \cref{sec_data_format_and_statistics}.
The dataset can be publicly accessed at Github: \href{https://github.com/yzhihao/MCSCSet}{\textcolor{blue}{https://github.com/yzhihao/MCSCSet}}.

\subsection{Medical Query Selection} \label{data_collection}
Building a large-scale Chinese spelling correction benchmark dataset in the medical domain is challenging. As demonstrated in the \cref{sec_introduction}, the first difficulty arises from the data collection. To ensure the practicality of medical-domain spelling corrector in real-world applications, the spelling errors should come from the real medical scenarios. To this end,
we collect a large-scale query log set $\mathcal{Q}$ with more than \textit{900k} samples from a real-world medical encyclopedia application named Tencent Yidian. Then, we remove queries that include personal information, such as name, ID number, and personal address. In addition, we further filter out queries with too long (more than 50 Chinese characters) or too short (less than 3), and queries without medical entities. Finally, we select about \textit{200k} queries containing common error-prone medical entities as a dataset to be annotated.

\subsection{Annotation Process} \label{annotation_process}

\begin{algorithm}[tb]
\caption{A high level description of the annotation procedure.}
\label{alg:Annotation_algorithm}
\textbf{Input}: A large scale query log set $\mathcal{Q}$ with more than $900,000$ samples. \\
\textbf{Output}: A medical query dataset $\mathcal{D}$ with $|\mathcal{D}|$ sample pairs.\\
\begin{algorithmic}[1] 
\STATE Clean the query set (e.g., removing queries with sensitive personal information and without medical entities, etc.)
\FOR {$t = 1 \to  |\mathcal{D}|$}
\STATE Initialize the flag $F \leftarrow False$
\STATE Expert annotators judge whether the raw sentence includes wrong entities and if so, assign flag $F \leftarrow True$, and correct the misspelled entities.\\
\IF{$F == False$}
\STATE Expert annotators annotate the corresponding erroneous entities.
\ENDIF
\STATE Expert annotators mark the error types and indexes of wrong medical entities, resulting the $t$-th annotated sample. 
\ENDFOR
\STATE \textbf{return}: The medical-domain Chinese spelling correction dataset (\ours).
\end{algorithmic}
\end{algorithm}

After going through medical query selection and getting a dataset to be annotated, we still have a second challenge to overcome. That is the annotation of misspelled medical entities requires the annotators to equip themselves with well-educated medical knowledge. In this regard, we hire annotators with medical backgrounds, such as medical students and hospital staff, to annotate medical entities of query and form a medical query error dataset $\mathcal{D}$ with $|\mathcal{D}| \approx 200,000$ samples. Figure \ref{fig:interface} shows the user interface of our annotation process, and the annotation steps are as follows:

\begin{itemize}
    \item \textbf{Step 1:} First, expert annotators find the medical entity in the query. Specifically, we first pre-label the medical entity with our medical entity recognition algorithm, and then the annotator needs to check the correctness of identified entities. After checking and correcting the entities identified by the algorithm, the annotator needs to mark the medical entities in the query with $\{\}$ to annotate the location.
    \item \textbf{Step 2:} Expert annotators check whether the query includes a wrong medical entity and correct the error. Expert annotators also mark the medical entity's error type. It should be noted that common words other than medical entities are not considered.
    \item \textbf{Step 3:} Suppose the raw query does not include a wrong entity. Expert annotators replace the correct medical entity with an erroneous entity, and mark the corresponding error type. 
    \item \textbf{Step 4:} Finally, The annotator puts the corresponding wrong query, correct query, and error type together to form an annotated sample.
\end{itemize}
The above annotation process is also formally summarized in algorithm \ref{alg:Annotation_algorithm}.

It should be noted that each sample is produced by two or three annotators. The first one annotates, and the second one checks. If disagreement occurs, the third annotator who is the most experienced would decide the final annotation.

\subsection{Data Format and Statistics} \label{sec_data_format_and_statistics}

\noindent \textbf{Schema of \ours:} 
The schema of \ours is shown in Table \ref{tab:sample_of_dataset}. Each sample includes five fields: wrong query, correct query, error location, error type, medical entity locations. Wrong query and correct query serve as input and output respectively in the medical-domain CSC task. In addition, exact error locations (\ie, indexes of wrong characters in the sequence) and error types are attached for each sample. They can provide rich information about spelling errors. Medical entities in queries are also marked because spelling errors are more likely to occur.

\begin{CJK*}{UTF8}{gkai}
\begin{table}[!t]
\caption{A sample from the \ours.}
\centering
\adjustbox{max width=1\linewidth}{
\setlength{\tabcolsep}{4mm}{
\begin{tabular}{ll}
\toprule
\textbf{Field}   & \textbf{Content}   \\
\midrule
Wrong query & 拨智尺的过程。\\
Correct query & 拔智齿的过程。\\
Error locations & (0, 2)   \\
Error types   & (visual, phonological) \\
Medical entity locations & ([0, 2]) \\
\bottomrule
\end{tabular}
}}
\label{tab:sample_of_dataset}
\end{table}
\end{CJK*}

\begin{table}[t]
\caption{Detailed statistics of \ours.}
\centering
\setlength{\tabcolsep}{4mm}{
\begin{tabular}{lr}
\toprule
\# Queries & 199,877 \\
Avg. query length & 10.90\\
Avg. \# misspelled characters per query & 1.86 \\
Avg. \# medical entities per query & 1.46 \\
\# Unique medical entities & 81,020 \\
\# Training samples & 157,194 \\
\# Validation samples & 19,652 \\
\# Test samples & 19,650 \\
\bottomrule
\end{tabular}}
\label{tab:statistics_of_dataset}
\end{table}

\begin{figure*}[!t]  
	\centering
	\subfigure[Distribution of query length.] {
	\includegraphics[width=0.3\linewidth,height=3.55cm]{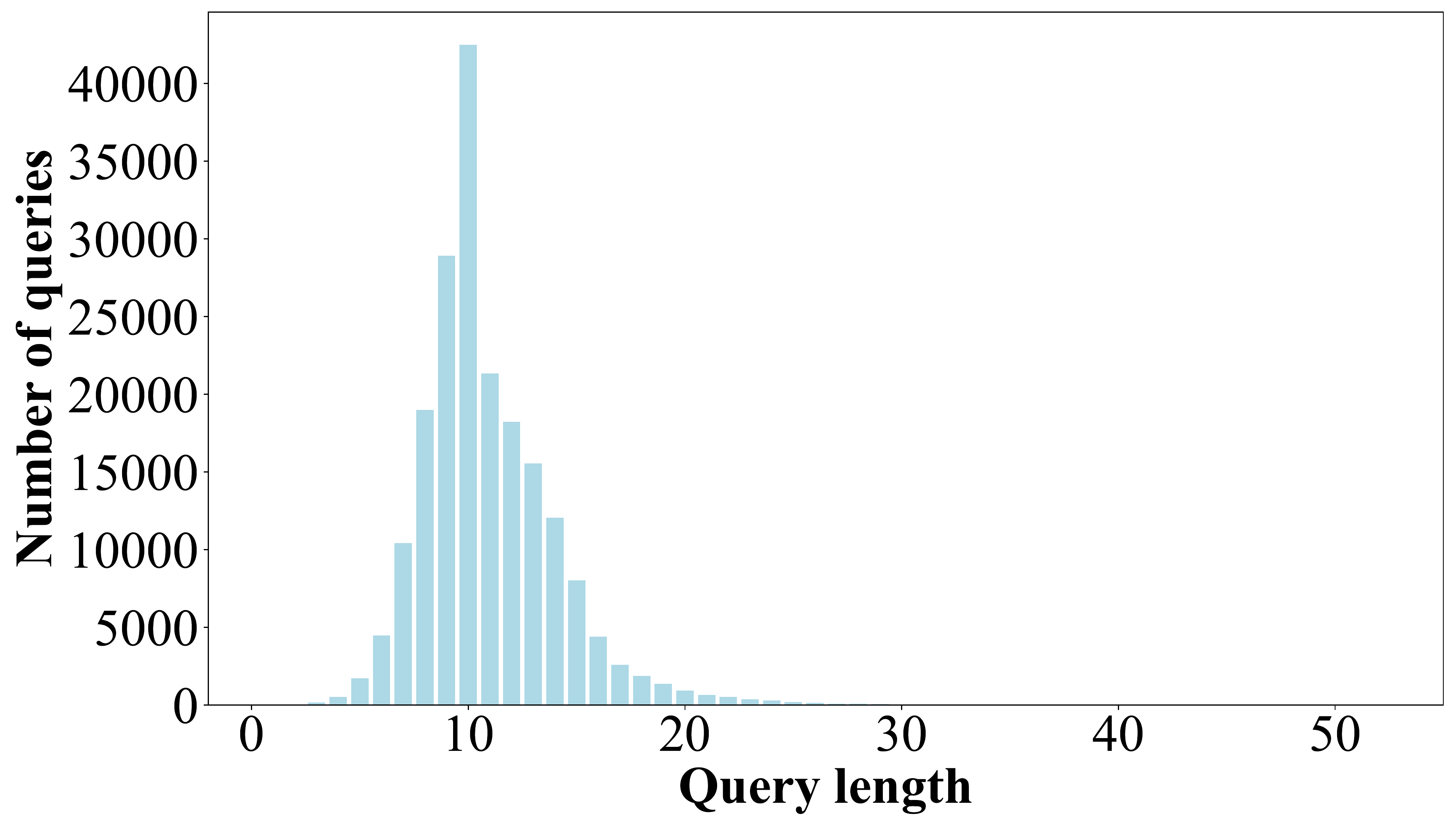}
	}	
	\subfigure[Distribution of medical entities' frequency.]{
	\includegraphics[width=0.3\linewidth,height=3.5cm]{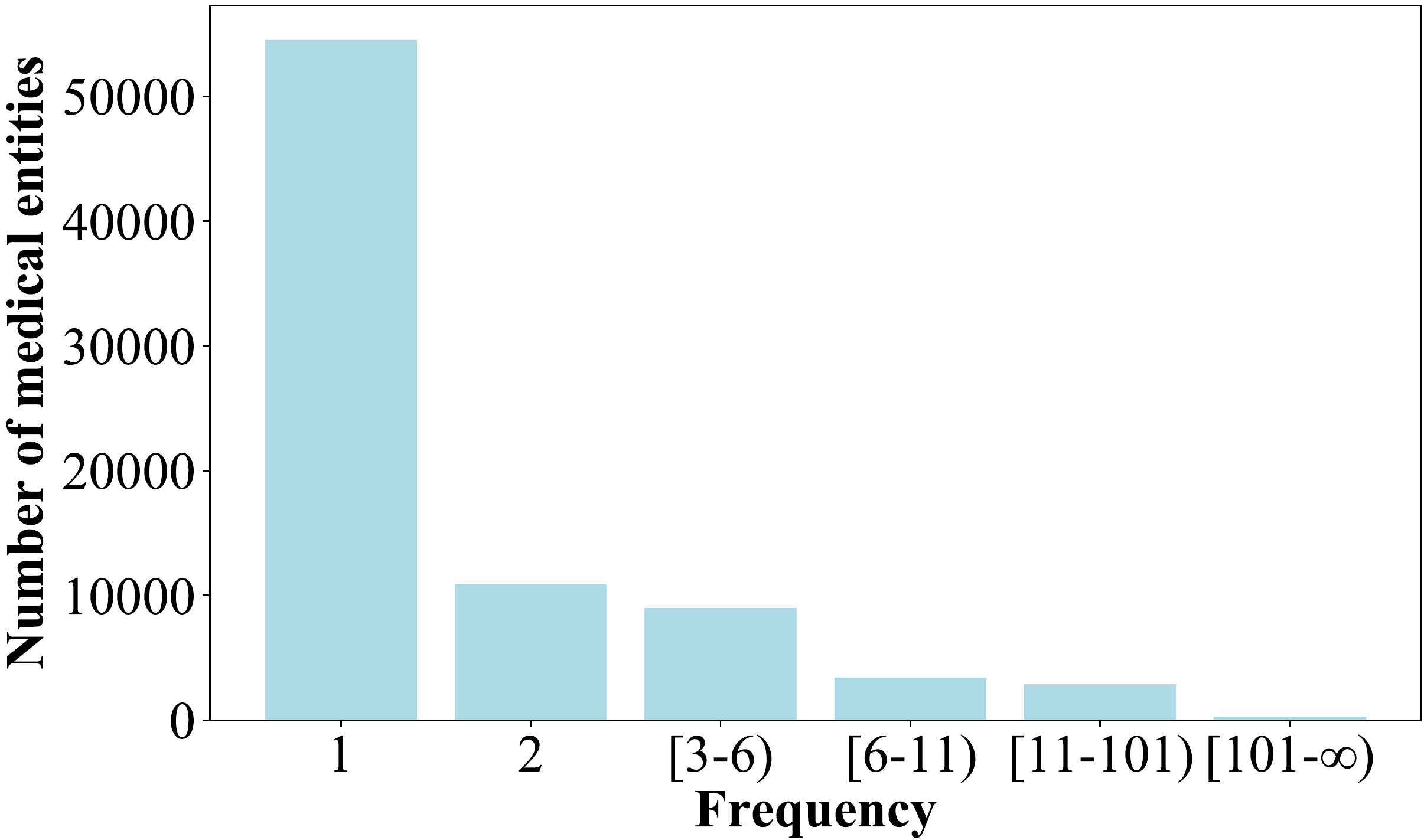}
	}	
	\subfigure[Distribution of entity error types.]{
	\includegraphics[width=0.3\linewidth,height=3.5cm ]{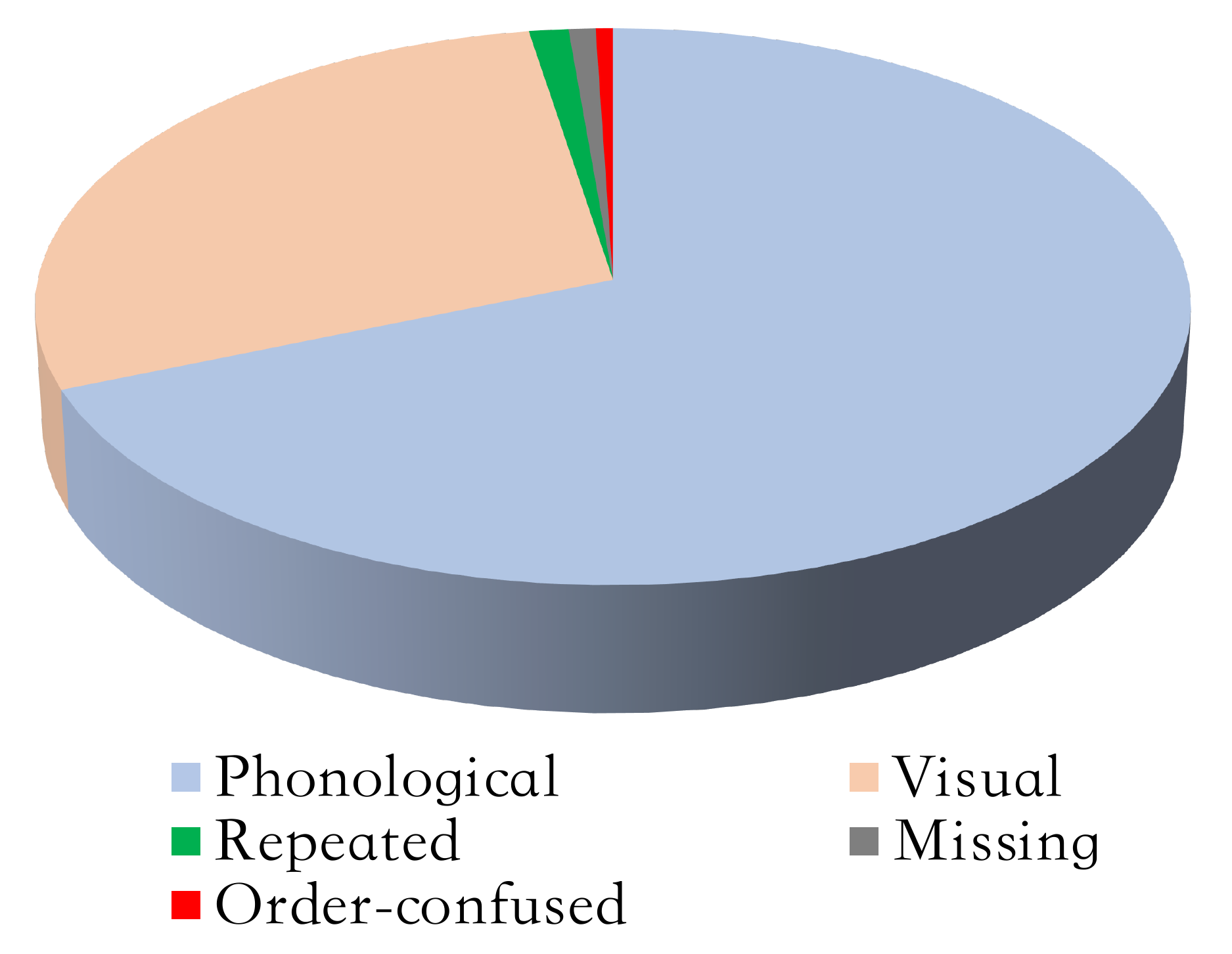}
	}	
	\caption{Visualization of the statistics for \ours.}
	\label{fig-Statistics}
\end{figure*}

\noindent \textbf{Statistics of \ours:} 
In Table \ref{tab:statistics_of_dataset}, we show the statistics of the \ours. We find the average query length of $10.9$ Chinese characters is short, and there are $1.86$ wrong characters per query, which may reduce the difficulty of spelling error correction. We attribute this to the inherent property of search queries because people tend to input relatively short text. The specific length distribution is demonstrated in Figure \ref{fig-Statistics} (a), and we can see that most medical queries contain less than twenty characters. Besides, there are an average of 1.46 medical entities in each query, and the total number of unique medical entities is up to $81,020$. Figure \ref{fig-Statistics} (b) shows the distribution of medical entities' frequency, and we can see that most of the medical entities appear less than or equal to five times in our dataset. Figure \ref{fig-Statistics} (c) shows the distribution of entity error types. From the figure, we can find that the error types of visual and phonological account for the highest proportion, about $96\%$ in total. Besides, our dataset also contains error types of order-confused, repeated, and missing. In other words, our dataset can also be used for syntax error correction.

\noindent \textbf{Medical Confusion Set:} 
On top of the \ours, we further construct a dictionary-like medical confusion set. Given a medical-domain error-prone character, we can easily find its corresponding common erroneous characters according to the medical confusion set. There are several public confusion sets in open-domain CSC, but they cannot be well applied in a specific domain like medicine. Therefore, we present the medical confusion set for the first time. Specifically, we collect all the spelling errors in the \ours and obtain abundant correct-erroneous character pairs. We find $2,623$ different characters are misspelled in our dataset by statistical analysis. Most of the characters have one to twenty corresponding erroneous characters. Several examples are shown in Table \ref{tab:sample_of_med_confusionset}. Additionally, we analyze high-frequency medical entities of all 81,020 that appear in our dataset, which means these entities are more prone to misspellings in real-world medical scenarios. As shown in Table \ref{high_frequency_entities}, we can find the top 5 high-frequency entities and the corresponding wrong entity sets from our medical confusion set.

\begin{CJK*}{UTF8}{gkai}
\begin{table}[!t]
\caption{Several examples in our medical confusion set.}
\centering
\adjustbox{max width=1.2\linewidth}{
\setlength{\tabcolsep}{1.4mm}{
\begin{tabular}{cl}
\toprule
\textbf{   Character   }   & \textbf{Corresponding misspelled characters}   \\
\midrule
疣 & 犹, 由, 庞, 忧, 莜, 旁, 优, 疮, 尤\\
酮 & 醇, 侗, 痛, 桐, 同, 彤, 铜, 筒, 通, 桶,\\
椎 & 谁, 柱, 锥, 追, 锤, 坠, 垂\\
氨 & 铵, 安, 胺, 按\\
呤 & 令,铃\\
\bottomrule
\end{tabular}
}}
\label{tab:sample_of_med_confusionset}
\end{table}
\end{CJK*}


\begin{CJK*}{UTF8}{gkai}
\begin{table}[!t]
\centering
\caption{Top 5 high-frequency entities.}
\setlength{\tabcolsep}{1mm}{
\begin{tabular}{lll}
\toprule
\textbf{Entities}   & \textbf{Frequency} & \textbf{Misspelled entities}   \\
\midrule
怀孕 \,(pregnancy) & 3,799 & (坏孕,还孕,淮孕)\\
发烧 \,(fever) & 2,483& (发稍,发梢,花烧)\\
咳嗽 \,(cough) & 2,060& (核嗽,咳嗖,咳嗦)\\
糖尿病 \,(diabetes) &1,890 & (唐尿病,糖尿崩,糖尿症)\\
感冒 \,(cold) &1,863 & (感昌,感帽,赶冒) \\
\bottomrule
\end{tabular}
}
\label{high_frequency_entities}
\end{table}
\end{CJK*}

\section{Experiments} \label{sec_experiment}
In this section, we conduct a series of studies on our \ours. 
First, we benchmark some typical CSC models on the \ours to establish baselines for future research. 
Then, we demonstrate the superiority of our dataset by comparing it with the automatically-generated dataset. 
Next, we show the effectiveness of our medical confusion set by comparing it with the open-domain confusion set. 
Finally, we investigate how the medical query's average number of misspelled characters affects the model performance.

\subsection{Experiment Setup}

\noindent \textbf{Experiment Configurations:} \textbf{(i) Problem Revisiting.} Chinese medical query spelling correction is to correct the misspelled characters in a medical query. In the experiment, the input is a medical query that possibly contains misspelled characters, and the output is a sentence with the same length as the input. For each character in the input sentence, if it is correct, the character in the same position in the output sentence should be unchanged; if it is misspelled, the character in the same position in the output sentence should be the corresponding correct character. \textbf{(ii) Dataset Filtering and Splitting.} 
There are five different types of errors in our dataset. 
Since we focus on the task of Chinese medical query spelling correction, we filtered out the medical queries containing errors other than spelling errors and obtained the remaining 196,496 samples.
For simplicity, we still call the remaining data \ours.
Finally, we split \ours into a training set (157,194), a validation set (19,652), and a test set (19,650) with a ratio of 8:1:1. \textbf{(iii) Samples that Need no Correction.} In real-world applications, only a small part of medical queries contains spelling errors, while all samples in \ours include spelling errors and need correction.
If we train the correction model with the original \ours, the model would be prone to believe that every input query must have spelling errors simply. 
As a result, the model would likely transform initially correct medical queries into erroneous ones.
Therefore, it is necessary to set a certain percentage of training samples to be in the form of correct-correct pair.
Meanwhile, the percentage should not be too small if the correction model cannot be efficiently trained.
In our actual practice, we set the percentage as 50\%, which is in line with \textit{SIGHAN-15}.



\noindent \textbf{Evaluation Metrics:}
Results are reported at the detection level and the correction level. 
A medical query is processed correctly at the detection level if and only if all spelling errors in the query are recognized, and it can be further considered to be processed successfully at the correction level if and only if all misspelled characters are replaced with right ones.
We report the precision (Prec.), recall (Rec.), and F1 scores on both levels.


\subsection{Benchmark Models}
\begin{figure}[t]
\begin{minipage}[t]{1.0\linewidth}
  \centering
  \centerline{\includegraphics[width=8.0cm]{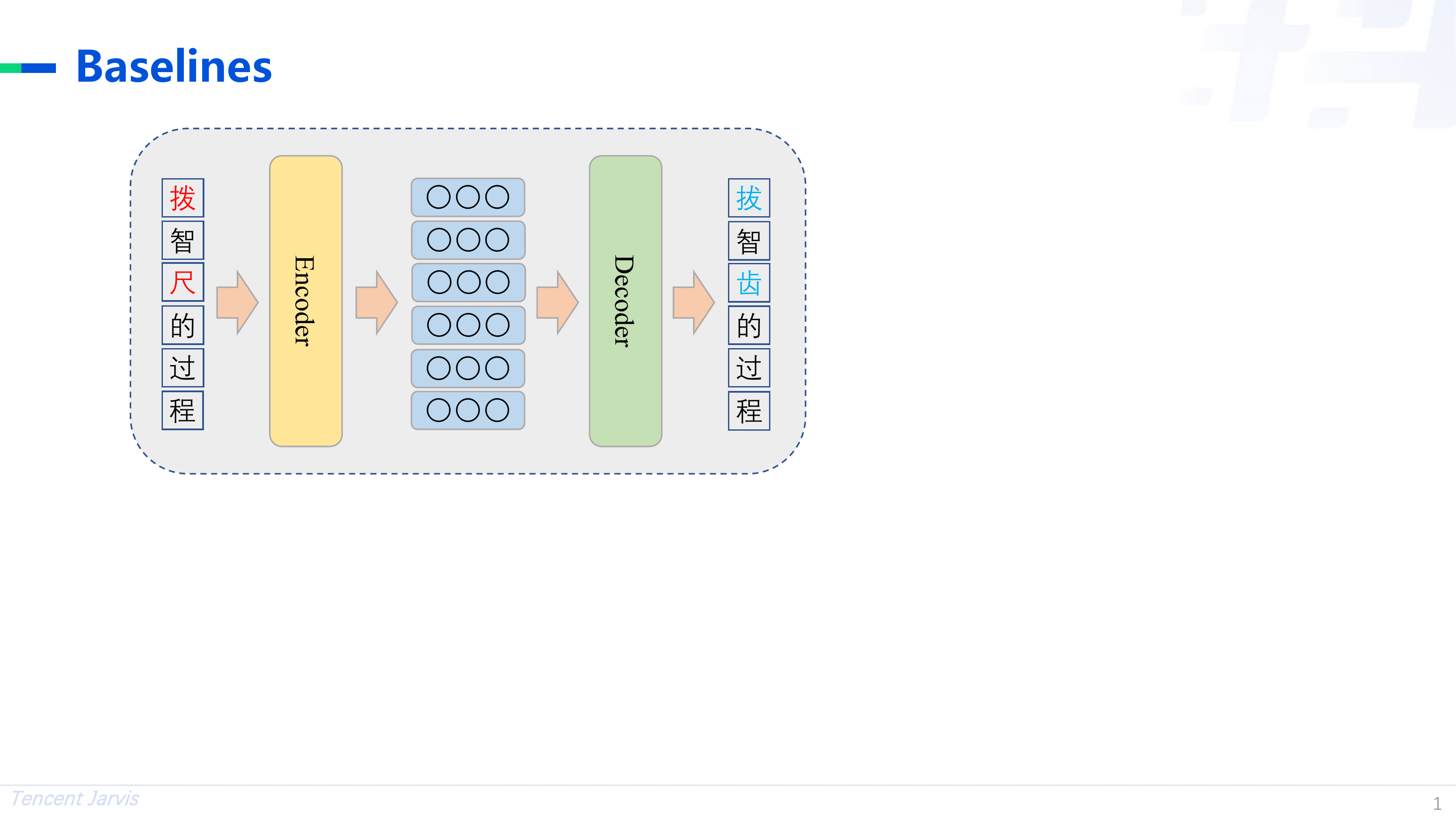}}
\end{minipage}
\caption{The general framework of existing CSC models.}
\label{fig:general_framework}
\end{figure}

The state-of-the-art methods in the \textit{SIGHAN} benchmark are almost based on pre-training CSC tasks on a large-scale automatically-generated dataset. To help understand the existing correction methods, we show the general framework in Figure \ref{fig:general_framework}. As depicted in the figure, the general framework is structured as an encoder-decoder architecture, in which the encoder is usually a pre-trained language model (\eg, BERT) and the decoder is typically a classifier (\eg, linear layer). Some of them have not released the code, while the others have not provided a pre-training dataset or the pre-trained weights, which makes their work hard to reproduce or apply in our \ours.
Therefore, we experiment with some representative and widely used spelling error correction methods as follows:

\noindent \textit{\textbf{BERT-Corrector}} \cite{li-etal-2021-exploration}: This method treats CSC as a non-autoregressive generation task and employs BERT as the model backbone. First, the input sentence is encoded by BERT to obtain the hidden representations of each character. Then, a classifier (\eg, linear layer) is used to pick the correction character from the whole vocabulary set for each character. According to our experiments, this is a simple but effective method that exceeds many methods with complicated designs.


\noindent \textit{\textbf{Soft-Masked BERT}} \cite{zhang2020spelling}: This method introduces the soft-masking strategy in BERT to improve error detection performance. Concretely, \textit{Soft-Masked BERT} first uses BERT as the encoder. Next, a Bi-GRU network is employed to detect the error probability for each character. The error probability is then used to weight the original and mask embedding to obtain the final character embedding. Last, a BERT-based network is applied to correct errors. With the error probability of character to incorporate the mask embedding, this method narrows the gap between CSC and the pretraining task of BERT (\ie, masked language model), thus achieving excellent results.

\noindent \textit{\textbf{MedBERT-Corrector}}: We build this model to take advantage of medical-domain knowledge. 
This method is similar to the above \textit{BERT-Corrector}. The only difference is that the encoder is replaced by PCL-MedBERT, \footnote{\href{https://code.ihub.org.cn/projects/1775}{https://code.ihub.org.cn/projects/1775}} which is a well-known pre-trained medical language model proposed by the Intelligent Medical Research Group at the Peng Cheng Laboratory, with excellent performance in medical question matching and named entity recognition.

\noindent \textbf{Implementation Details:}
The experiments are conducted with Pytorch \cite{paszke2019pytorch} and HuggingFace Transformers \cite{wolf2019huggingface}.
For all methods based on the pre-trained language model, we set different learning rates for the pre-trained language model and the subsequent classifier as this shows better performance in our experiments. 
We search the learning rate of the pre-trained language model in [$1\times 10^{-5}$, $3\times 10^{-5}$,$5\times 10^{-5}$, $1\times 10^{-4}$], the best is $5\times 10^{-5}$.
Similarly, the best learning rate of the classifier is $3\times 10^{-4}$ in [$1\times 10^{-4}$, $3\times 10^{-4}$, $5\times 10^{-4}$, $1\times 10^{-3}$].
The warmup and dropout ratio are both 0.1.
We train the model for a total of 20 epochs with a batch size of 64, and early stopping is activated when the validation scores do not improve for 5 epochs.
Especially for \textit{Soft-Masked BERT}, we adopt the hyper-parameters proposed in \cite{zhang2020spelling} with the Bi-GRU, the Bi-GRU hidden unit of 256 and the balance weight $\lambda$ in the objective function of 0.8.
All the reported results are averaged over 5 runs with different random seeds.
The experiments are conducted on NVIDIA Tesla V100 with 32GB memory.

\subsection{Benchmark Experiments}

\noindent \textbf{Main Results}:
To bulid a valid benchmark and facilitate future research on our task, we experiment with the above representative CSC methods on our \ours.
Table \ref{tab:main_exp} shows the main results, and we have the following observations:  
(1) Methods based on pre-trained language models achieve promising results on our \ours. 
Specifically, they all have correction-level F1 score over 80\%.
The potential reasons are twofold: \textit{i)} pre-trained language models can offer general linguistic knowledge, which is fundamental for spelling error correction. \textit{ii)} \ours provides the domain knowledge as well as correction supervision, which makes the correction models well generalizable in the medical domain.
(2) \textit{MedBERT-Corrector} outperforms \textit{BERT-Corrector} by a bit of margin, which indicates the effectiveness of incorporating external medical knowledge to a certain extent. We should also notice that the effectiveness is limited for the pre-trained medical language model, which enlightens the need for other valid methods of introducing medical knowledge. Incorporating a medical knowledge graph during training probably works.
(3) \textit{Soft-Masked BERT} obtains the best results on our dataset. We attribute this to the specially-designed soft-masking strategy to narrow the gap between spelling error correction and masked language model. By incorporating the mask embedding, the misspelled characters in medical entities are more accurately located.

In general, pre-trained language models have been proved to be highly effective in our task. They can improve the benchmark performance by effectively incorporating medical knowledge and considering more characteristics of spelling errors in the medical domain.

\begin{table*}[!t]
\centering
\caption{Performances of benchmark models on \ours. The best results are highlighted in bold.}
\setlength{\tabcolsep}{6 mm}{
\begin{tabular}{c|ccc|ccc}
\toprule
\multicolumn{1}{c|}{\multirow{2}{*}{Method}}   & \multicolumn{3}{c|}{Detection-level} & \multicolumn{3}{c}{Correction-level}   \\
\cline { 2 - 7 } & Prec. (\%) & Rec. (\%) & F1 (\%) & Prec. (\%) & Rec. (\%) & F1 (\%) \\
\midrule
\textit{BERT-Corrector} \cite{li-etal-2021-exploration} & \textbf{87.05} & 86.08 & 86.55 & 80.93 & 80.05 & 80.49  \\
\textit{MedBERT-Corrector} & 87.01 & 86.25 & 86.63 & 80.98 & 80.24 & 80.61  \\
\textit{Soft-Masked BERT} \cite{zhang2020spelling} & 87.03 & \textbf{86.29} & \textbf{86.66} & \textbf{81.22} & \textbf{80.54} & \textbf{80.88}  \\
\bottomrule
\end{tabular}
}
\label{tab:main_exp}
\end{table*}

\noindent \textbf{Comparison between \ours and Automatically-generated Dataset}:
To demonstrate the necessity and great value of constructing the dataset with specialist annotation, we compare our \ours with the automatically-generated dataset.
The automatically-generated dataset is commonly used in open-domain CSC for the following reasons.
First, existing open-domain CSC datasets are limited in scale incapable of training satisfactory CSC models.
Second, the automatically-generated data is easy to acquire, and the dataset scale can be massive.
Third, the quality of the automatically-generated data is acceptable, which can be used to pre-train the CSC models.
Automatically generating a training sample (wrong-correct pair) is straightforward.
According to the confusion set, the wrong query can be obtained by replacing some characters in the correct query with the corresponding erroneous characters.
In this way, large-scale open-domain datasets can be generated automatically, and they are used to pre-train CSC models followed by fine-tuning on the target dataset. 
Although automatically-generated datasets have achieved open-domain CSC, we argue this paradigm is not suitable for CSC in the medical domain as the open-domain confusion set is inapplicable in the medical domain. 
To validate this, we first construct the automatically-generated dataset (\textit{AGSet-Open}) based on the training set and validation set of \ours with a commonly used open-domain confusion set \cite{wang2019confusionset}. 
For a fair comparison, we set the average number of misspelled characters to 1.8 to keep consistency to the original \ours. 
Specifically, we train \textit{BERT-Corrector} on \textit{AGSet-Open} and \ours respectively and evaluate them on the same test set of \ours. 
As the results shown in Table \ref{tab:auto_confusion_dataset}, we can observe that the correction-level F1 score of the model trained on \textit{AGSet-Open} is under 40\%, which is even lower than half of that on \ours.
We conjecture this is because the automatically-generated dataset has poor quality and fails to simulate the practical distribution of Chinese medical query spelling correction.

\begin{table*}[!t]
\centering
\caption{Performances of \textit{BERT-Corrector} trained on the automatically-generated datasets (\textit{AGSet-Open}, \textit{AGSet-Med}) and \ours, respectively. Notably, \textit{AGSet-Open} is constructed based on \ours and open-domain confusion set~\cite{wang2019confusionset}, while \textit{AGSet-Med} is built based on the \ours and our medical confusion set.}
\setlength{\tabcolsep}{7 mm}{
\begin{tabular}{c|ccc|ccc}
\toprule
\multicolumn{1}{c|}{\multirow{2}{*}{Dataset}}   & \multicolumn{3}{c|}{Detection-level} & \multicolumn{3}{c}{Correction-level}   \\
\cline { 2 - 7 } & Prec. (\%) & Rec. (\%) & F1 (\%) & Prec. (\%) & Rec. (\%) & F1 (\%) \\
\midrule
\ours & 87.05 & 86.08 & 86.55 & 80.93 & 80.05 & 80.49  \\
\textit{AGSet-Open} & 62.77 & 52.69 & 57.29 & 43.61 & 36.62 & 39.81  \\
\textit{AGSet-Med} & 72.25 & 61.32 & 66.34 & 62.24 & 55.16 & 60.19  \\

\bottomrule
\end{tabular}
}
\label{tab:auto_confusion_dataset}
\end{table*}

\noindent \textbf{Effectiveness of the Medical Confusion Set}:
In this part, we compare our medical confusion set with the open-domain confusion set to illustrate the effectiveness and superiority of our medical confusion set.
As mentioned in the last subsection, the confusion set can be used to generate training samples automatically.
However, results show that the open-domain confusion set is not suitable for the medical domain.
Compared with the open-domain confusion set, our medical confusion set is constructed on top of \ours.
Therefore, it is customized for spelling corrections in the medical domain, which is of high quality and can be used to further automatically generate training samples that are suitable for Chinese medical query spelling correction. 
To prove the above claims, we construct \textit{AGSet-Med} based on \ours and our medical confusion set in a way similar to the construction of \textit{AGSet-Open}
The only difference is we replace the original open-domain confusion set with our medical confusion set. 
For a fair comparison, we also set the average number of wrong characters as 1.86. 
Finally, we train \textit{BERT-Corrector} on \textit{AGSet-Med} and report the evaluation results in Table \ref{tab:auto_confusion_dataset}. 
We can see that the model trained on \textit{AGSet-Med} outperforms the one trained on \textit{AGSet-Open}, which indicates that \textit{AGSet-Med} has better quality than \textit{AGSet-Open} and further demonstrates the effectiveness and superiority of our medical confusion set.
The medical confusion set is of great value for medical-domain CSC. First, it is customized for medical-domain CSC and contains the misspelling feature of medical texts since it is constructed on top of \ours. Second, the medical confusion set can be employed to help the candidate characters filtering. Specifically, once a character is detected misspelled, to determine the target corrected character from the whole vocabulary, we can set larger weights for the erroneous corresponding characters in the medical confusion set. Third, the medical confusion set can be used to automatically generate large-scale CSC data in the medical domain. Therefore, it is promising to obtain a superior correction model by first pre-training it on the generated large-scale medical-domain CSC data following fine-tuning it on the specific target dataset.

\noindent \textbf{Impact of the Average Number of Misspelled Characters of the Medical Query}:
We study how the average number of misspelled characters affects the model performance.
Intuitively, a medical query with more erroneous characters is more challenging to correct because the number of misspelled characters measures the correction difficulty to a certain extent.
To obtain \ours with the different average number of misspelled characters, we employ our medical confusion set to add erroneous characters on top of the original \ours.
Specifically, for each correct character in the medical entities of a medical query, with a certain probability, we change it to one of its corresponding erroneous characters according to the medical confusion set.
We can obtain \ours with a different average number of misspelled characters by controlling the probability. 
Then we train \textit{BERT-Corrector} on these datasets respectively and report their performances (F1 score) in Figure \ref{fig:misspelled characters}. 
As shown in the figure, the model performance deteriorates as the average number of misspelled characters increases. 
Besides, when the average number of misspelled characters exceeds about 2.6, the model performance degrades relatively faster.
We conjecture the reason is that after certain point, the model has less context to correct misspelled characters.

\section{Conclusion And Future Work} \label{sec_conclusion}
In this paper, we proposed the medical-domain Chinese spelling correction task and introduced \ours, a large-scale dataset dedicated to this task, annotated by medical specialists. 
On top of \ours, we also proposed the medical confusion set, which could facilitate the automatic data generation process.
We conducted a series of experiments to demonstrate the necessity and effectiveness of \ours and the medical confusion set. Moreover, we benchmarked several representative spelling correction methods to provide baselines for future research. As shown by our experiments, the current spelling correction algorithms still have much room for improvement in the medical domain, which means there still lies space to delve deeper for future work.

\newpage
\bibliographystyle{ACM-Reference-Format}
\bibliography{sample-base}


\begin{thebibliography}{43}


\ifx \showCODEN    \undefined \def \showCODEN     #1{\unskip}     \fi
\ifx \showDOI      \undefined \def \showDOI       #1{#1}\fi
\ifx \showISBNx    \undefined \def \showISBNx     #1{\unskip}     \fi
\ifx \showISBNxiii \undefined \def \showISBNxiii  #1{\unskip}     \fi
\ifx \showISSN     \undefined \def \showISSN      #1{\unskip}     \fi
\ifx \showLCCN     \undefined \def \showLCCN      #1{\unskip}     \fi
\ifx \shownote     \undefined \def \shownote      #1{#1}          \fi
\ifx \showarticletitle \undefined \def \showarticletitle #1{#1}   \fi
\ifx \showURL      \undefined \def \showURL       {\relax}        \fi
\providecommand\bibfield[2]{#2}
\providecommand\bibinfo[2]{#2}
\providecommand\natexlab[1]{#1}
\providecommand\showeprint[2][]{arXiv:#2}

\bibitem[Afli et~al\mbox{.}(2016)]%
        {afli2016using}
\bibfield{author}{\bibinfo{person}{Haithem Afli}, \bibinfo{person}{Zhengwei
  Qiu}, \bibinfo{person}{Andy Way}, {and} \bibinfo{person}{P{\'a}raic
  Sheridan}.} \bibinfo{year}{2016}\natexlab{}.
\newblock \showarticletitle{Using SMT for OCR error correction of historical
  texts}. In \bibinfo{booktitle}{\emph{Proceedings of the Tenth International
  Conference on Language Resources and Evaluation}}. \bibinfo{pages}{962--966}.
\newblock


\bibitem[Arslan et~al\mbox{.}(2021)]%
        {arslan2021detecting}
\bibfield{author}{\bibinfo{person}{Recep~Sinan Arslan},
  \bibinfo{person}{Necaattin Barisci}, \bibinfo{person}{Nursal Arici}, {and}
  \bibinfo{person}{Sabri Kocer}.} \bibinfo{year}{2021}\natexlab{}.
\newblock \showarticletitle{Detecting and correcting automatic speech
  recognition errors with a new model}.
\newblock \bibinfo{journal}{\emph{Turkish Journal of Electrical Engineering \&
  Computer Sciences}} \bibinfo{volume}{29}, \bibinfo{number}{5}
  (\bibinfo{year}{2021}), \bibinfo{pages}{2298--2311}.
\newblock


\bibitem[Cheng et~al\mbox{.}(2020)]%
        {cheng2020spellgcn}
\bibfield{author}{\bibinfo{person}{Xingyi Cheng}, \bibinfo{person}{Weidi Xu},
  \bibinfo{person}{Kunlong Chen}, \bibinfo{person}{Shaohua Jiang},
  \bibinfo{person}{Feng Wang}, \bibinfo{person}{Taifeng Wang},
  \bibinfo{person}{Wei Chu}, {and} \bibinfo{person}{Yuan Qi}.}
  \bibinfo{year}{2020}\natexlab{}.
\newblock \showarticletitle{SpellGCN: Incorporating phonological and visual
  similarities into language models for Chinese spelling check}. In
  \bibinfo{booktitle}{\emph{Proceedings of the 58th Annual Meeting of the
  Association for Computational Linguistics}}. \bibinfo{pages}{871--881}.
\newblock


\bibitem[Chung et~al\mbox{.}(2014)]%
        {gru}
\bibfield{author}{\bibinfo{person}{Junyoung Chung}, \bibinfo{person}{Caglar
  Gulcehre}, \bibinfo{person}{Kyunghyun Cho}, {and} \bibinfo{person}{Yoshua
  Bengio}.} \bibinfo{year}{2014}\natexlab{}.
\newblock \showarticletitle{Empirical evaluation of gated recurrent neural
  networks on sequence modeling}. In \bibinfo{booktitle}{\emph{NIPS Workshop on
  Deep Learning}}.
\newblock


\bibitem[Devlin et~al\mbox{.}(2019)]%
        {devlin-etal-2019-bert}
\bibfield{author}{\bibinfo{person}{Jacob Devlin}, \bibinfo{person}{Ming-Wei
  Chang}, \bibinfo{person}{Kenton Lee}, {and} \bibinfo{person}{Kristina
  Toutanova}.} \bibinfo{year}{2019}\natexlab{}.
\newblock \showarticletitle{{BERT}: Pre-training of deep bidirectional
  transformers for language understanding}. In
  \bibinfo{booktitle}{\emph{Proceedings of the 2019 Conference of the North
  {A}merican Chapter of the Association for Computational Linguistics: Human
  Language Technologies, Volume 1 (Long and Short Papers)}}.
  \bibinfo{pages}{4171--4186}.
\newblock


\bibitem[Drobac and Lind{\'e}n(2020)]%
        {drobac2020optical}
\bibfield{author}{\bibinfo{person}{Senka Drobac} {and} \bibinfo{person}{Krister
  Lind{\'e}n}.} \bibinfo{year}{2020}\natexlab{}.
\newblock \showarticletitle{Optical character recognition with neural networks
  and post-correction with finite state methods}.
\newblock \bibinfo{journal}{\emph{International Journal on Document Analysis
  and Recognition}} \bibinfo{volume}{23}, \bibinfo{number}{4}
  (\bibinfo{year}{2020}), \bibinfo{pages}{279--295}.
\newblock


\bibitem[Duan et~al\mbox{.}(2018)]%
        {duan2018error}
\bibfield{author}{\bibinfo{person}{Jianyong Duan}, \bibinfo{person}{Tianxiao
  Ji}, {and} \bibinfo{person}{Hao Wang}.} \bibinfo{year}{2018}\natexlab{}.
\newblock \showarticletitle{Error correction for search engine by mining bad
  case}.
\newblock \bibinfo{journal}{\emph{IEICE transactions on Information and
  Systems}} \bibinfo{volume}{101}, \bibinfo{number}{7} (\bibinfo{year}{2018}),
  \bibinfo{pages}{1938--1945}.
\newblock


\bibitem[Gao et~al\mbox{.}(2010)]%
        {gao2010large}
\bibfield{author}{\bibinfo{person}{Jianfeng Gao}, \bibinfo{person}{Chris
  Quirk}, {et~al\mbox{.}}} \bibinfo{year}{2010}\natexlab{}.
\newblock \showarticletitle{A large scale ranker-based system for search query
  spelling correction}. In \bibinfo{booktitle}{\emph{The 23rd International
  Conference on Computational Linguistics}}.
\newblock


\bibitem[Guo et~al\mbox{.}(2019)]%
        {guo2019spelling}
\bibfield{author}{\bibinfo{person}{Jinxi Guo}, \bibinfo{person}{Tara~N
  Sainath}, {and} \bibinfo{person}{Ron~J Weiss}.}
  \bibinfo{year}{2019}\natexlab{}.
\newblock \showarticletitle{A spelling correction model for end-to-end speech
  recognition}. In \bibinfo{booktitle}{\emph{IEEE International Conference on
  Acoustics, Speech and Signal Processing}}. IEEE, \bibinfo{pages}{5651--5655}.
\newblock


\bibitem[Guo et~al\mbox{.}(2021)]%
        {guo2021global}
\bibfield{author}{\bibinfo{person}{Zhao Guo}, \bibinfo{person}{Yuan Ni},
  \bibinfo{person}{Keqiang Wang}, \bibinfo{person}{Wei Zhu}, {and}
  \bibinfo{person}{Guotong Xie}.} \bibinfo{year}{2021}\natexlab{}.
\newblock \showarticletitle{Global attention decoder for chinese spelling error
  correction}. In \bibinfo{booktitle}{\emph{Findings of the Association for
  Computational Linguistics}}. \bibinfo{pages}{1419--1428}.
\newblock


\bibitem[Hong et~al\mbox{.}(2019)]%
        {hong2019faspell}
\bibfield{author}{\bibinfo{person}{Yuzhong Hong}, \bibinfo{person}{Xianguo Yu},
  \bibinfo{person}{Neng He}, \bibinfo{person}{Nan Liu}, {and}
  \bibinfo{person}{Junhui Liu}.} \bibinfo{year}{2019}\natexlab{}.
\newblock \showarticletitle{FASPell: A fast, adaptable, simple, powerful
  Chinese spell checker based on DAE-decoder paradigm}. In
  \bibinfo{booktitle}{\emph{Proceedings of the 5th Workshop on Noisy
  User-generated Text}}. \bibinfo{pages}{160--169}.
\newblock


\bibitem[Huang et~al\mbox{.}(2021)]%
        {huang2021phmospell}
\bibfield{author}{\bibinfo{person}{Li Huang}, \bibinfo{person}{Junjie Li},
  \bibinfo{person}{Weiwei Jiang}, \bibinfo{person}{Zhiyu Zhang},
  \bibinfo{person}{Minchuan Chen}, \bibinfo{person}{Shaojun Wang}, {and}
  \bibinfo{person}{Jing Xiao}.} \bibinfo{year}{2021}\natexlab{}.
\newblock \showarticletitle{PHMOSpell: Phonological and morphological knowledge
  guided Chinese spelling check}. In \bibinfo{booktitle}{\emph{Proceedings of
  the 59th Annual Meeting of the Association for Computational Linguistics and
  the 11th International Joint Conference on Natural Language Processing
  (Volume 1: Long Papers)}}. \bibinfo{pages}{5958--5967}.
\newblock


\bibitem[Ji et~al\mbox{.}(2021)]%
        {ji2021spellbert}
\bibfield{author}{\bibinfo{person}{Tuo Ji}, \bibinfo{person}{Hang Yan}, {and}
  \bibinfo{person}{Xipeng Qiu}.} \bibinfo{year}{2021}\natexlab{}.
\newblock \showarticletitle{SpellBERT: A lightweight pretrained model for
  Chinese spelling check}. In \bibinfo{booktitle}{\emph{Proceedings of the 2021
  Conference on Empirical Methods in Natural Language Processing}}.
  \bibinfo{pages}{3544--3551}.
\newblock


\bibitem[Kipf and Welling(2017)]%
        {KipfW2017gcn}
\bibfield{author}{\bibinfo{person}{Thomas~N. Kipf} {and} \bibinfo{person}{Max
  Welling}.} \bibinfo{year}{2017}\natexlab{}.
\newblock \showarticletitle{Semi-supervised classification with graph
  convolutional networks}. In \bibinfo{booktitle}{\emph{5th International
  Conference on Learning Representations}}.
\newblock


\bibitem[Li et~al\mbox{.}(2021)]%
        {li-etal-2021-exploration}
\bibfield{author}{\bibinfo{person}{Chong Li}, \bibinfo{person}{Cenyuan Zhang},
  \bibinfo{person}{Xiaoqing Zheng}, {and} \bibinfo{person}{Xuanjing Huang}.}
  \bibinfo{year}{2021}\natexlab{}.
\newblock \showarticletitle{Exploration and Exploitation: Two Ways to Improve
  {C}hinese Spelling Correction Models}. In
  \bibinfo{booktitle}{\emph{Proceedings of the 59th Annual Meeting of the
  Association for Computational Linguistics and the 11th International Joint
  Conference on Natural Language Processing (Volume 2: Short Papers)}}.
  \bibinfo{pages}{441--446}.
\newblock


\bibitem[Lin and Chen(2015)]%
        {lin2015ntou}
\bibfield{author}{\bibinfo{person}{Chuan-Jie Lin} {and}
  \bibinfo{person}{Shao-Heng Chen}.} \bibinfo{year}{2015}\natexlab{}.
\newblock \showarticletitle{NTOU Chinese grammar checker for CGED shared task}.
  In \bibinfo{booktitle}{\emph{Proceedings of the 2nd Workshop on Natural
  Language Processing Techniques for Educational Applications}}.
  \bibinfo{pages}{15--19}.
\newblock


\bibitem[Liu et~al\mbox{.}(2021)]%
        {liu2021plome}
\bibfield{author}{\bibinfo{person}{Shulin Liu}, \bibinfo{person}{Tao Yang},
  \bibinfo{person}{Tianchi Yue}, \bibinfo{person}{Feng Zhang}, {and}
  \bibinfo{person}{Di Wang}.} \bibinfo{year}{2021}\natexlab{}.
\newblock \showarticletitle{PLOME: Pre-training with misspelled knowledge for
  Chinese spelling correction}. In \bibinfo{booktitle}{\emph{Proceedings of the
  59th Annual Meeting of the Association for Computational Linguistics and the
  11th International Joint Conference on Natural Language Processing (Volume 1:
  Long Papers)}}. \bibinfo{pages}{2991--3000}.
\newblock


\bibitem[Liu et~al\mbox{.}(2013)]%
        {liu2013hybrid}
\bibfield{author}{\bibinfo{person}{Xiaodong Liu}, \bibinfo{person}{Kevin
  Cheng}, \bibinfo{person}{Yanyan Luo}, \bibinfo{person}{Kevin Duh}, {and}
  \bibinfo{person}{Yuji Matsumoto}.} \bibinfo{year}{2013}\natexlab{}.
\newblock \showarticletitle{A hybrid Chinese spelling correction using language
  model and statistical machine translation with reranking}. In
  \bibinfo{booktitle}{\emph{Proceedings of the Seventh SIGHAN Workshop on
  Chinese Language Processing}}. \bibinfo{pages}{54--58}.
\newblock


\bibitem[Martins and Silva(2004)]%
        {martins2004spelling}
\bibfield{author}{\bibinfo{person}{Bruno Martins} {and}
  \bibinfo{person}{M{\'a}rio~J Silva}.} \bibinfo{year}{2004}\natexlab{}.
\newblock \showarticletitle{Spelling correction for search engine queries}. In
  \bibinfo{booktitle}{\emph{International Conference on Natural Language
  Processing}}.
\newblock


\bibitem[Mokhtar et~al\mbox{.}(2018)]%
        {mokhtar2018ocr}
\bibfield{author}{\bibinfo{person}{Kareem Mokhtar}, \bibinfo{person}{Syed~Saqib
  Bukhari}, {and} \bibinfo{person}{Andreas Dengel}.}
  \bibinfo{year}{2018}\natexlab{}.
\newblock \showarticletitle{OCR error correction: State-of-the-art vs an
  NMT-based approach}. In \bibinfo{booktitle}{\emph{13th IAPR International
  Workshop on Document Analysis Systems}}. IEEE, \bibinfo{pages}{429--434}.
\newblock


\bibitem[Moon and Burstein(2007)]%
        {moon2007ontology}
\bibfield{author}{\bibinfo{person}{Jane Moon} {and} \bibinfo{person}{Frada
  Burstein}.} \bibinfo{year}{2007}\natexlab{}.
\newblock \showarticletitle{Ontology-based spelling correction for searching
  medical information}.
\newblock \bibinfo{journal}{\emph{Semantic Web Technologies and E-business:
  Toward the Integrated Virtual Organization and Business Process Automation}}
  (\bibinfo{year}{2007}), \bibinfo{pages}{384--404}.
\newblock


\bibitem[Paszke et~al\mbox{.}(2019)]%
        {paszke2019pytorch}
\bibfield{author}{\bibinfo{person}{Adam Paszke}, \bibinfo{person}{Sam Gross},
  \bibinfo{person}{Francisco Massa}, \bibinfo{person}{Adam Lerer},
  \bibinfo{person}{James Bradbury}, \bibinfo{person}{Gregory Chanan},
  \bibinfo{person}{Trevor Killeen}, \bibinfo{person}{Zeming Lin},
  \bibinfo{person}{Natalia Gimelshein}, \bibinfo{person}{Luca Antiga},
  {et~al\mbox{.}}} \bibinfo{year}{2019}\natexlab{}.
\newblock \showarticletitle{Pytorch: An imperative style, high-performance deep
  learning library}.
\newblock \bibinfo{journal}{\emph{Advances in Neural Information Processing
  Systems}} (\bibinfo{year}{2019}).
\newblock


\bibitem[Qiu et~al\mbox{.}(2020)]%
        {qiu2020pre}
\bibfield{author}{\bibinfo{person}{Xipeng Qiu}, \bibinfo{person}{Tianxiang
  Sun}, \bibinfo{person}{Yige Xu}, \bibinfo{person}{Yunfan Shao},
  \bibinfo{person}{Ning Dai}, {and} \bibinfo{person}{Xuanjing Huang}.}
  \bibinfo{year}{2020}\natexlab{}.
\newblock \showarticletitle{Pre-trained models for natural language processing:
  A survey}.
\newblock \bibinfo{journal}{\emph{Science China Technological Sciences}}
  \bibinfo{volume}{63}, \bibinfo{number}{10} (\bibinfo{year}{2020}),
  \bibinfo{pages}{1872--1897}.
\newblock


\bibitem[Sch{\"u}tze et~al\mbox{.}(2008)]%
        {schutze2008introduction}
\bibfield{author}{\bibinfo{person}{Hinrich Sch{\"u}tze},
  \bibinfo{person}{Christopher~D Manning}, {and} \bibinfo{person}{Prabhakar
  Raghavan}.} \bibinfo{year}{2008}\natexlab{}.
\newblock \bibinfo{booktitle}{\emph{Introduction to information retrieval}}.
  Vol.~\bibinfo{volume}{39}.
\newblock \bibinfo{publisher}{Cambridge University Press Cambridge}.
\newblock


\bibitem[Tseng et~al\mbox{.}(2015a)]%
        {tseng2015introduction}
\bibfield{author}{\bibinfo{person}{Yuen-Hsien Tseng}, \bibinfo{person}{Lung-Hao
  Lee}, \bibinfo{person}{Li-Ping Chang}, {and} \bibinfo{person}{Hsin-Hsi
  Chen}.} \bibinfo{year}{2015}\natexlab{a}.
\newblock \showarticletitle{Introduction to SIGHAN 2015 bake-off for Chinese
  spelling check}. In \bibinfo{booktitle}{\emph{Proceedings of the Eighth
  SIGHAN Workshop on Chinese Language Processing}}. \bibinfo{pages}{32--37}.
\newblock


\bibitem[Tseng et~al\mbox{.}(2015b)]%
        {tseng-etal-2015-introduction}
\bibfield{author}{\bibinfo{person}{Yuen-Hsien Tseng}, \bibinfo{person}{Lung-Hao
  Lee}, \bibinfo{person}{Li-Ping Chang}, {and} \bibinfo{person}{Hsin-Hsi
  Chen}.} \bibinfo{year}{2015}\natexlab{b}.
\newblock \showarticletitle{Introduction to {SIGHAN} 2015 Bake-off for
  {C}hinese spelling check}. In \bibinfo{booktitle}{\emph{Proceedings of the
  Eighth {SIGHAN} Workshop on {C}hinese Language Processing}}.
  \bibinfo{publisher}{Association for Computational Linguistics},
  \bibinfo{address}{Beijing, China}, \bibinfo{pages}{32--37}.
\newblock


\bibitem[Wang et~al\mbox{.}(2018)]%
        {wang2018hybrid}
\bibfield{author}{\bibinfo{person}{Dingmin Wang}, \bibinfo{person}{Yan Song},
  \bibinfo{person}{Jing Li}, \bibinfo{person}{Jialong Han}, {and}
  \bibinfo{person}{Haisong Zhang}.} \bibinfo{year}{2018}\natexlab{}.
\newblock \showarticletitle{A hybrid approach to automatic corpus generation
  for Chinese spelling check}. In \bibinfo{booktitle}{\emph{Proceedings of the
  2018 Conference on Empirical Methods in Natural Language Processing}}.
\newblock


\bibitem[Wang et~al\mbox{.}(2019)]%
        {wang2019confusionset}
\bibfield{author}{\bibinfo{person}{Dingmin Wang}, \bibinfo{person}{Yi Tay},
  {and} \bibinfo{person}{Li Zhong}.} \bibinfo{year}{2019}\natexlab{}.
\newblock \showarticletitle{Confusionset-guided pointer networks for chinese
  spelling check}. In \bibinfo{booktitle}{\emph{Proceedings of the 2019
  Conference on Association for Computational Linguistics}}.
\newblock


\bibitem[Wolf et~al\mbox{.}(2019)]%
        {wolf2019huggingface}
\bibfield{author}{\bibinfo{person}{Thomas Wolf}, \bibinfo{person}{Lysandre
  Debut}, \bibinfo{person}{Victor Sanh}, \bibinfo{person}{Julien Chaumond},
  \bibinfo{person}{Clement Delangue}, \bibinfo{person}{Anthony Moi},
  \bibinfo{person}{Pierric Cistac}, \bibinfo{person}{Tim Rault},
  \bibinfo{person}{R{\'e}mi Louf}, \bibinfo{person}{Morgan Funtowicz},
  {et~al\mbox{.}}} \bibinfo{year}{2019}\natexlab{}.
\newblock \showarticletitle{Huggingface's Transformers: State-of-the-art
  natural language processing}.
\newblock \bibinfo{journal}{\emph{arXiv preprint arXiv:1910.03771}}
  (\bibinfo{year}{2019}).
\newblock


\bibitem[Wu et~al\mbox{.}(2013a)]%
        {wu2013chinese}
\bibfield{author}{\bibinfo{person}{Shih-Hung Wu}, \bibinfo{person}{Chao-Lin
  Liu}, {and} \bibinfo{person}{Lung-Hao Lee}.}
  \bibinfo{year}{2013}\natexlab{a}.
\newblock \showarticletitle{Chinese Spelling Check Evaluation at SIGHAN
  Bake-off 2013.}. In \bibinfo{booktitle}{\emph{SIGHAN@ IJCNLP}}. Citeseer,
  \bibinfo{pages}{35--42}.
\newblock


\bibitem[Wu et~al\mbox{.}(2013b)]%
        {wu-etal-2013-chinese}
\bibfield{author}{\bibinfo{person}{Shih-Hung Wu}, \bibinfo{person}{Chao-Lin
  Liu}, {and} \bibinfo{person}{Lung-Hao Lee}.}
  \bibinfo{year}{2013}\natexlab{b}.
\newblock \showarticletitle{{C}hinese Spelling Check Evaluation at {SIGHAN}
  Bake-off 2013}. In \bibinfo{booktitle}{\emph{Proceedings of the Seventh
  {SIGHAN} Workshop on {C}hinese Language Processing}}.
  \bibinfo{publisher}{Asian Federation of Natural Language Processing},
  \bibinfo{address}{Nagoya, Japan}, \bibinfo{pages}{35--42}.
\newblock


\bibitem[Xie et~al\mbox{.}(2015)]%
        {xie2015chinese}
\bibfield{author}{\bibinfo{person}{Weijian Xie}, \bibinfo{person}{Peijie
  Huang}, \bibinfo{person}{Xinrui Zhang}, \bibinfo{person}{Kaiduo Hong},
  \bibinfo{person}{Qiang Huang}, \bibinfo{person}{Bingzhou Chen}, {and}
  \bibinfo{person}{Lei Huang}.} \bibinfo{year}{2015}\natexlab{}.
\newblock \showarticletitle{Chinese spelling check system based on n-gram
  model}. In \bibinfo{booktitle}{\emph{Proceedings of the Eighth SIGHAN
  Workshop on Chinese Language Processing}}. \bibinfo{pages}{128--136}.
\newblock


\bibitem[Xu et~al\mbox{.}(2021)]%
        {xu-etal-2021-read}
\bibfield{author}{\bibinfo{person}{Heng-Da Xu}, \bibinfo{person}{Zhongli Li},
  \bibinfo{person}{Qingyu Zhou}, \bibinfo{person}{Chao Li},
  \bibinfo{person}{Zizhen Wang}, \bibinfo{person}{Yunbo Cao},
  \bibinfo{person}{Heyan Huang}, {and} \bibinfo{person}{Xian-Ling Mao}.}
  \bibinfo{year}{2021}\natexlab{}.
\newblock \showarticletitle{Read, listen, and see: leveraging multimodal
  information helps {C}hinese spell checking}. In
  \bibinfo{booktitle}{\emph{Findings of the Association for Computational
  Linguistics}}. \bibinfo{publisher}{Association for Computational
  Linguistics}, \bibinfo{address}{Online}, \bibinfo{pages}{716--728}.
\newblock


\bibitem[Yang et~al\mbox{.}(2019)]%
        {yang2019post}
\bibfield{author}{\bibinfo{person}{Li Yang}, \bibinfo{person}{Ying Li},
  \bibinfo{person}{Jin Wang}, {and} \bibinfo{person}{Zhuo Tang}.}
  \bibinfo{year}{2019}\natexlab{}.
\newblock \showarticletitle{Post text processing of Chinese speech recognition
  based on bidirectional LSTM networks and CRF}.
\newblock \bibinfo{journal}{\emph{Electronics}} \bibinfo{volume}{8},
  \bibinfo{number}{11} (\bibinfo{year}{2019}), \bibinfo{pages}{1248}.
\newblock


\bibitem[Yeh et~al\mbox{.}(2015)]%
        {yeh2015chinese}
\bibfield{author}{\bibinfo{person}{Jui-Feng Yeh}, \bibinfo{person}{Wen-Yi
  Chen}, {and} \bibinfo{person}{Mao-Chuan Su}.}
  \bibinfo{year}{2015}\natexlab{}.
\newblock \showarticletitle{Chinese spelling checker based on an inverted index
  list with a rescoring mechanism}.
\newblock \bibinfo{journal}{\emph{ACM Transactions on Asian and Low-Resource
  Language Information Processing}} \bibinfo{volume}{14}, \bibinfo{number}{4}
  (\bibinfo{year}{2015}), \bibinfo{pages}{1--28}.
\newblock


\bibitem[Yeh et~al\mbox{.}(2014)]%
        {yeh2014chinese}
\bibfield{author}{\bibinfo{person}{Jui-Feng Yeh}, \bibinfo{person}{Yun-Yun Lu},
  \bibinfo{person}{Chen-Hsien Lee}, \bibinfo{person}{Yu-Hsiang Yu}, {and}
  \bibinfo{person}{Yong-Ting Chen}.} \bibinfo{year}{2014}\natexlab{}.
\newblock \showarticletitle{Chinese word spelling correction based on rule
  induction}. In \bibinfo{booktitle}{\emph{Proceedings of The Third CIPS-SIGHAN
  Joint Conference on Chinese Language Processing}}. \bibinfo{pages}{139--145}.
\newblock


\bibitem[Yu and Li(2014)]%
        {yu2014chinese}
\bibfield{author}{\bibinfo{person}{Junjie Yu} {and} \bibinfo{person}{Zhenghua
  Li}.} \bibinfo{year}{2014}\natexlab{}.
\newblock \showarticletitle{Chinese spelling error detection and correction
  based on language model, pronunciation, and shape}. In
  \bibinfo{booktitle}{\emph{Proceedings of The Third CIPS-SIGHAN Joint
  Conference on Chinese Language Processing}}. \bibinfo{pages}{220--223}.
\newblock


\bibitem[Yu et~al\mbox{.}(2014)]%
        {yu-etal-2014-overview}
\bibfield{author}{\bibinfo{person}{Liang-Chih Yu}, \bibinfo{person}{Lung-Hao
  Lee}, \bibinfo{person}{Yuen-Hsien Tseng}, {and} \bibinfo{person}{Hsin-Hsi
  Chen}.} \bibinfo{year}{2014}\natexlab{}.
\newblock \showarticletitle{Overview of {SIGHAN} 2014 Bake-off for {C}hinese
  Spelling Check}. In \bibinfo{booktitle}{\emph{Proceedings of The Third
  {CIPS}-{SIGHAN} Joint Conference on {C}hinese Language Processing}}.
  \bibinfo{publisher}{Association for Computational Linguistics},
  \bibinfo{address}{Wuhan, China}, \bibinfo{pages}{126--132}.
\newblock


\bibitem[Zhang et~al\mbox{.}(2021)]%
        {zhang2021correcting}
\bibfield{author}{\bibinfo{person}{Ruiqing Zhang}, \bibinfo{person}{Chao Pang},
  \bibinfo{person}{Chuanqiang Zhang}, \bibinfo{person}{Shuohuan Wang},
  \bibinfo{person}{Zhongjun He}, \bibinfo{person}{Yu Sun}, \bibinfo{person}{Hua
  Wu}, {and} \bibinfo{person}{Haifeng Wang}.} \bibinfo{year}{2021}\natexlab{}.
\newblock \showarticletitle{Correcting Chinese spelling errors with phonetic
  pre-training}. In \bibinfo{booktitle}{\emph{Findings of the Association for
  Computational Linguistics}}. \bibinfo{pages}{2250--2261}.
\newblock


\bibitem[Zhang et~al\mbox{.}(2020)]%
        {zhang2020spelling}
\bibfield{author}{\bibinfo{person}{Shaohua Zhang}, \bibinfo{person}{Haoran
  Huang}, \bibinfo{person}{Jicong Liu}, {and} \bibinfo{person}{Hang Li}.}
  \bibinfo{year}{2020}\natexlab{}.
\newblock \showarticletitle{Spelling error correction with soft-masked BERT}.
  In \bibinfo{booktitle}{\emph{Proceedings of the 58th Annual Meeting of the
  Association for Computational Linguistics}}. \bibinfo{pages}{882--890}.
\newblock


\bibitem[Zhang et~al\mbox{.}(2019)]%
        {zhang2019investigation}
\bibfield{author}{\bibinfo{person}{Shiliang Zhang}, \bibinfo{person}{Ming Lei},
  {and} \bibinfo{person}{Zhijie Yan}.} \bibinfo{year}{2019}\natexlab{}.
\newblock \showarticletitle{Investigation of Transformer based spelling
  correction model for CTC-based end-to-end mandarin speech recognition.}. In
  \bibinfo{booktitle}{\emph{Interspeech}}. \bibinfo{pages}{2180--2184}.
\newblock


\bibitem[Zhao et~al\mbox{.}(2017)]%
        {zhao2017hybrid}
\bibfield{author}{\bibinfo{person}{Hai Zhao}, \bibinfo{person}{Deng Cai},
  \bibinfo{person}{Yang Xin}, \bibinfo{person}{Yuzhu Wang}, {and}
  \bibinfo{person}{Zhongye Jia}.} \bibinfo{year}{2017}\natexlab{}.
\newblock \showarticletitle{A hybrid model for Chinese spelling check}.
\newblock \bibinfo{journal}{\emph{ACM Transactions on Asian and Low-Resource
  Language Information Processing}} \bibinfo{volume}{16}, \bibinfo{number}{3}
  (\bibinfo{year}{2017}), \bibinfo{pages}{1--22}.
\newblock


\bibitem[Zhou et~al\mbox{.}(2015)]%
        {zhou2015context}
\bibfield{author}{\bibinfo{person}{Xiaofang Zhou}, \bibinfo{person}{An Zheng},
  \bibinfo{person}{Jiaheng Yin}, \bibinfo{person}{Rudan Chen},
  \bibinfo{person}{Xianyang Zhao}, \bibinfo{person}{Wei Xu},
  \bibinfo{person}{Wenqing Cheng}, \bibinfo{person}{Tian Xia},
  \bibinfo{person}{Simon Lin}, {et~al\mbox{.}}}
  \bibinfo{year}{2015}\natexlab{}.
\newblock \showarticletitle{Context-sensitive spelling correction of
  consumer-generated content on health care}.
\newblock \bibinfo{journal}{\emph{JMIR Medical Informatics}}
  \bibinfo{volume}{3}, \bibinfo{number}{3} (\bibinfo{year}{2015}),
  \bibinfo{pages}{e4211}.
\newblock


\end{thebibliography}
\end{document}